\documentclass[fleqn,10pt]{wlscirep}
\usepackage[utf8]{inputenc}
\usepackage[T1]{fontenc}

\usepackage{graphicx}
\usepackage{tikz}
\usepackage{setspace}
\usepackage{longtable}
\usepackage{array}
\usepackage{subcaption}
\usepackage{graphicx}

\title{Uncovering the Corona Virus Map \\ Using Deep Entities and Relationship Models}

\author[1]{Kuldeep Singh}
\author[1]{Puneet Singla} 
\author[1]{Ketan Sarode} 
\author[1]{Anurag Chandrakar}
\author[1,*]{Chetan Nichkawde}
\affil[1]{Innoplexus AG, Pune, India}
\affil[*]{Corresponding author: Dr.~Chetan Nichkawde, Head of Artificial Intelligence}

\newcommand{\tikzcircle}[2][red,fill=red]{\tikz[baseline=-0.5ex]\draw[#1,radius=#2] (0,0) circle ;}%

\begin{abstract}
We extract entities and relationships related to COVID-19 from a corpus of articles related to Corona virus by employing a novel entities and relationship model. The entity recognition 
and relationship discovery models are trained with a multi-task learning objective 
on a large annotated corpus.
We employ a concept masking paradigm to prevent the evolution of neural networks functioning as an associative memory and induce right inductive bias guiding the network to make inference using only the context.
We uncover several import subnetworks, highlight important terms and concepts and elucidate
several treatment modalities employed in related ailments in the past.
\end{abstract}

\begin{document}

\flushbottom
\maketitle

\thispagestyle{empty}

\section{Introduction}
The recent outbreak of SARS-CoV-2 has led to a global pandemic with the total number of infections exceeding 6 million with more than 370000 mortality already.
The disease has been code named COVID-19 and had far reaching repercussions the world over. This article aims to uncover
the life science universe of the Corona virus and related ailments by employing some of the state-of-the-art natural language processing technologies applied to biomedical domain.
We took the corpus of about 40000 titles and abstracts released as a part of CORD-19 Open Research Challenge and applied our entity recognition and
relationship discovery models to construct a knowledge graph related to COVID-19.
In the process, we uncovered about 40000 entities and 80000 relationships.

This article presents our salient findings and is organized as follows.
Section \ref{sec:MEM} briefly describes our masked entities model and masked relationship model.
Section \ref{sec:coronanetwork} presents a network analysis of the knowledge network discovered by mining CORD-19 dataset.
The coverage of CORD-19 dataset may be not exhaustive and up-to-date. We took snapshot around April 15, 2020. Nevertheless, the primary aim of this work is to demonstrate the application of artificial
intelligence on condensing unstructured information in the biomedical domain to a sufficiently low entropy state so that some important leads can be established.

\section{\label{sec:MEM}The Entities and Relationship Model}
\begin{figure}
    \centering
    \includegraphics[scale=0.25]{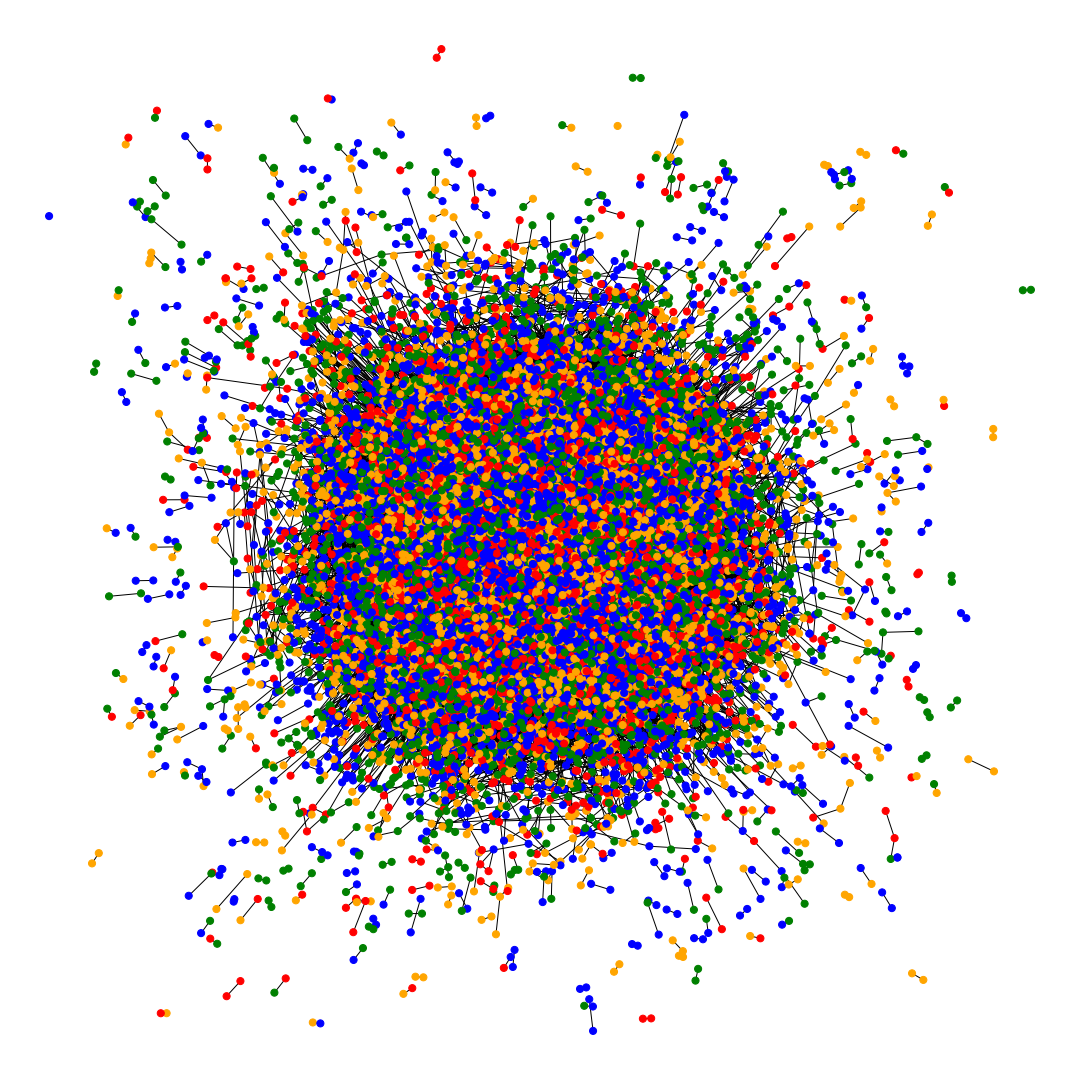}
    \caption{Full COVID-19 entities and relationship network. 
    \tikzcircle[blue,fill=blue]{4pt} -- Protein; \tikzcircle[green,fill=green]{4pt} -- Drug;
    \tikzcircle[red,fill=red]{4pt} -- Disease; \tikzcircle[orange,fill=orange]{4pt} -- Taxonomy}
    \label{fig:covidgr}
\end{figure}
We ran our entity recognition model that was trained on about 1 billion data points that we have built in-house.
A corpus of about 33 million titles and abstracts was tagged for 4 different kinds of entities -- protein, drug, disease, and taxonomy.
We employ a novel concept masking paradigm where the term occurrences were replaced by a dummy token.
Thus, essentially we remove the entire vocabulary associated with these entities from our training corpus.
This inductive bias guides the model to make inference using the surrounding context alone and helps us achieve state-of-the-art results on biomedical entity recognition problem\cite{nichkawde1}. 
We use a transformer architecture\cite{vaswani2017attention} with the following three joint end-to-end multitask learning objectives: 1) masked token prediction 2) next sentence prediction 3) entity type prediction.
We train the word piece tokenization algorithm on our in-house corpus to generate a vocabulary set of 30000 word pieces.
The network had 8 encoder layers with each layer composed of self-attention followed by a feedforward network.
Positional encodings were used in the beginning.
We will have an elaborate publication on this at a later date\cite{nichkawde1}.

We further uncover relationships between the entities by employing our relationship discovery model.
The entities are once again masked to guide the network to make an inference using the context alone.
The transformer architecture is once again used for encoding sentences. We use a novel bilinear
attention at the output to model interaction between contextualized embeddings of the two entities between whom we are trying to establish a relationship\cite{nichkawde2}.

\section{\label{sec:coronanetwork}The Corona virus network analysis}
We present here some of the important subnetworks and concepts uncovered as result of probing of CORD-19 dataset\cite{wang2020cord} by our neural network models. The findings in the paper are only suggestive and
the purpose of this work is to demonstrate the application of artificial intelligence to uncovering
important concepts and relationships in the life sciences domain. We hope to bring few import leads in sharp focus and help a researcher to narrow down the scope of his search for most important concepts and relationships. 
Figure \ref{fig:covidgr} shows the full set of all entities and relationships. We computed Katz centrality measure for each node in the network.
The Katz centrality measure can be understood follows: let $A$ be an $n \times n$ adjacency matrix of the
network with the element $A_{ij}$ being 1 if there is a relationship between node $i$ and node $j$
and zero otherwise. The powers of $A$ such as $k^{th}$ power $A^k$ is representative of paths
between two nodes through intermediaries.
The Katz centrality measure for the node $i$ defined as:
\[
C_{Katz}(i) = \sum_{k=1}^{\infty}\sum_{j=1}^{n} \alpha^k \left(A^k\right)_{ji}
\]
The attenuation factor $\alpha$ is chosen such that it is smaller than the reciprocal of the absolute value of the largest eigenvalue of A.
The top 20 concepts in the literature ranked by the normalized Katz centrality measure\cite{katz1953new} is shown in Table~\ref{tab:topconcepts}(a).
The top ranked concept is ofcourse coronavirus followed by cytokine. Concepts like spike protein and ACE2 also figure in the top 20. Several different kinds of viruses such as adenovirus and rotavirus also figure in the top 20 showing the diversity of the CORD-19 dataset.
\begin{table}[h!]
\scalebox{0.8}{
\begin{subtable}{0.7\linewidth}
\centering
\caption{Top concepts in the network}
 \begin{tabular}{|| c | c | c | c ||} 
 \hline
 Rank & Entity & Type & Centrality Measure \\
 \hline\hline
1 & coronavirus & taxonomy & 0.1050 \\
\hline
2 & cytokine & protein & 0.0696 \\
\hline
3 & pneumonia & disease & 0.0621 \\
\hline
4 & spike protein & protein & 0.0595 \\
\hline
5 & sars & disease & 0.0580 \\
\hline
6 & nucleocapsid protein & protein & 0.0506 \\
\hline
7 & ace2 & protein & 0.0450 \\
\hline
8 & type i ifn & drug & 0.0401 \\
\hline
9 & sars-cov & taxonomy & 0.0399 \\
\hline
10 & il-6 & protein & 0.0356 \\
\hline
11 & asthma & disease & 0.0347 \\
\hline
12 & adenovirus & taxonomy & 0.0344 \\
\hline
13 & pedv & taxonomy & 0.0339 \\
\hline
14 & respiratory syncytial virus & taxonomy & 0.0319 \\
\hline
15 & piglet & taxonomy & 0.0305 \\
\hline
16 & tnf-$\alpha$ & protein & 0.0301 \\
\hline
17 & ifn & protein & 0.0301 \\
\hline
18 & influenza virus & taxonomy & 0.0294 \\
\hline
19 & sars-cov-2 & taxonomy & 0.0292 \\
\hline
20 & covid-19 & disease & 0.0289 \\
\hline
\end{tabular}
\end{subtable}
}
\scalebox{0.8}{
\begin{subtable}{0.5\linewidth}
\centering
\caption{Top proteins in the network}
\begin{tabular}{|| c | c | c ||}
\hline
 Rank & Protein & Centrality Measure \\
 \hline\hline
\hline
1 & cytokine & 0.0696 \\
\hline
2 & spike protein & 0.0595 \\
\hline
3 & nucleocapsid protein & 0.0506 \\
\hline
4 & ace2 & 0.0450 \\
\hline
5 & il-6 & 0.0356 \\
\hline
6 & tnf-$\alpha$ & 0.0301 \\
\hline
7 & ifn & 0.0301 \\
\hline
8 & lectin & 0.0285 \\
\hline
9 & ifn-$\gamma$ & 0.0262 \\
\hline
10 & il-10 & 0.0262 \\
\hline
11 & nf-kb & 0.0260 \\
\hline
12 & envelope protein & 0.0252 \\
\hline
13 & il-8 & 0.0217 \\
\hline
14 & membrane protein & 0.0207 \\
\hline
15 & il-1$\beta$ & 0.0194 \\
\hline
16 & ifn-$\beta$ & 0.0187 \\
\hline
17 & tlr3 & 0.0181 \\
\hline
18 & irf3 & 0.0177 \\
\hline
19 & cxcl10 & 0.0174 \\
\hline
20 & ifitm3 & 0.0165 \\
\hline
 \end{tabular}
\end{subtable}
}
\caption{Concepts ranked by Katz centrality measure}
\label{tab:topconcepts}
\end{table}
\begin{figure}[h!]
    \centering
    \includegraphics[width=0.95\textwidth]{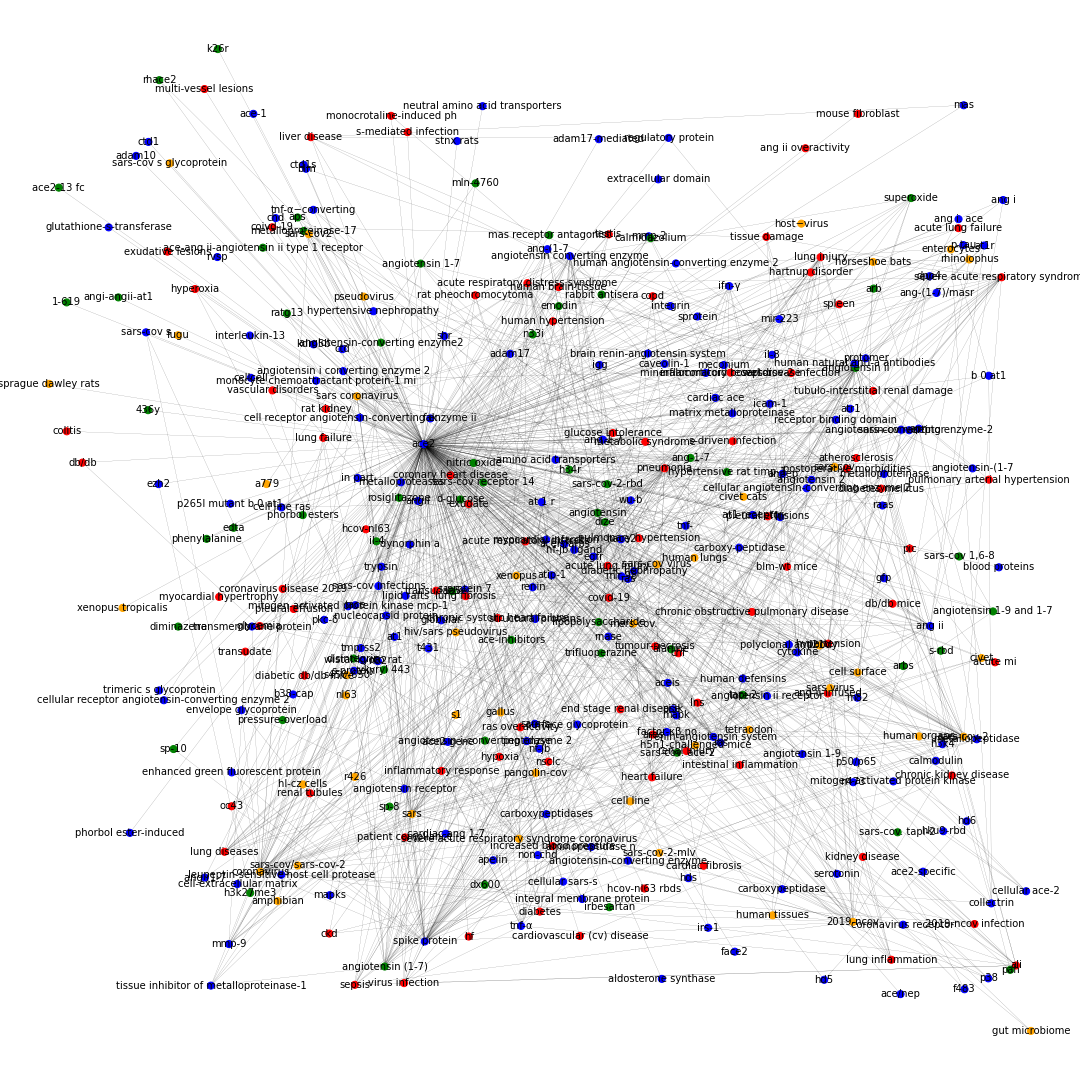}
    \caption{ACE2 subnetwork
        \tikzcircle[blue,fill=blue]{4pt} -- Protein; \tikzcircle[green,fill=green]{4pt} -- Drug;
        \tikzcircle[red,fill=red]{4pt} -- Disease; \tikzcircle[orange,fill=orange]{4pt} -- Taxonomy}
    \label{fig:ace2subnetwork}
\end{figure}
The top proteins are shown in Table~\ref{tab:topconcepts}(b). The adverse cytokine response by the immune system and cytokine storm occurring at a late stage of the
ailment is the top mentioned protein concept in the literature. Spike protein which binds so efficiently to the ACE2 receptor resulting in viral entry is ranked
second among all proteins. Spike protein is at the center of many therapeutic concepts under evaluation for possible treatments. Surprisingly, \emph{nucleocapsid protein} or \emph{N protein} is the third most important
protein concept discussed in the literature. \emph{N protein} is a multifunctional protein playing a pivotal role in enhancing the efficiency of viral transcription and assembly. Infact, the machinery related to \emph{N protein} was proposed as one of the top therapeutic target
in a recent work. It was proposed that targeting of viral translation by interfering with the eIF4F complex formation or the interactions between viral proteins N, Nsp2, Nsp8, and the transnational machinery may have therapeutic benefits\cite{gordon2020sars}.

\subsection{The ACE2 and Spike protein network}
\begin{figure}[h!]
    \centering
    \includegraphics[scale=0.25]{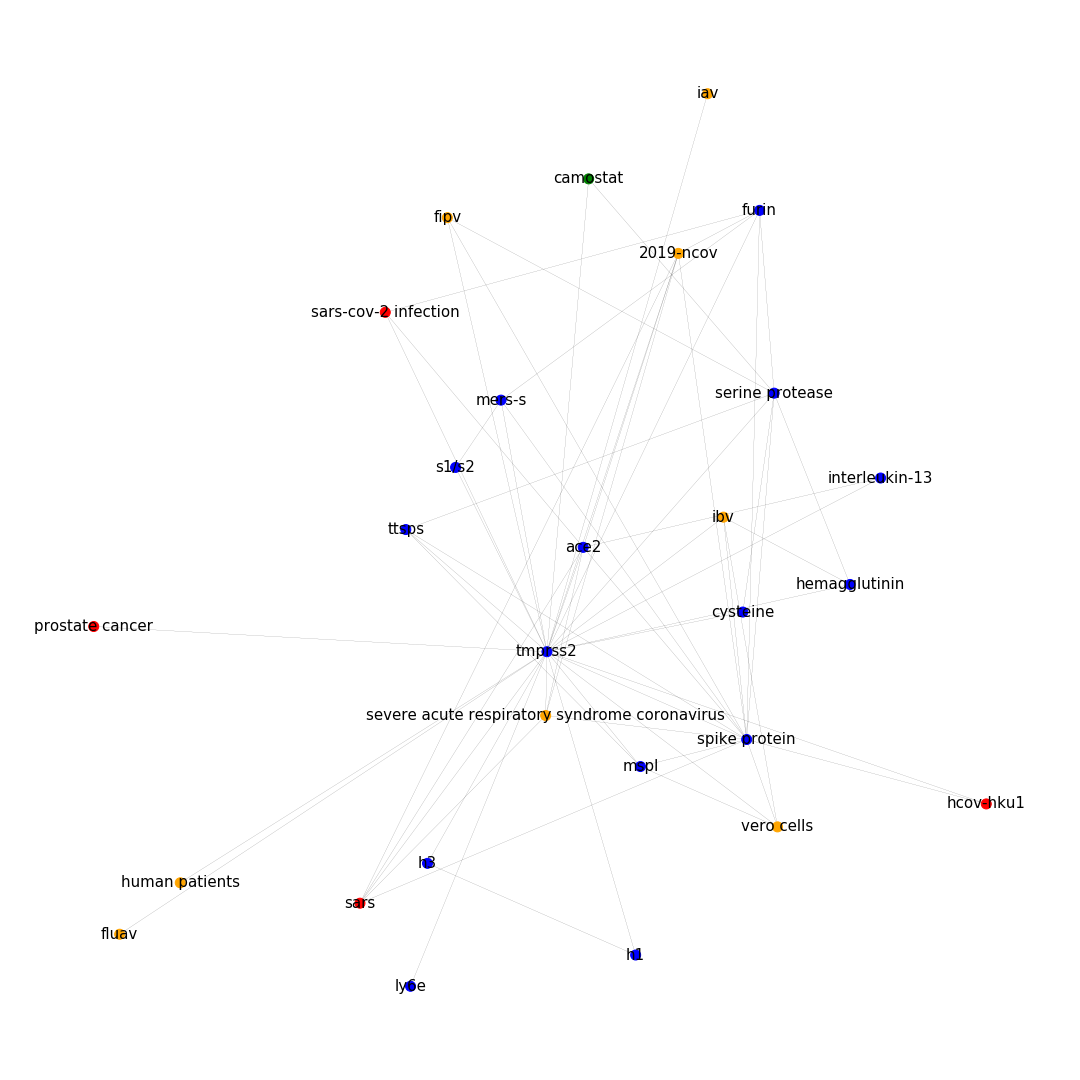}
    \caption{TMPRSS2 network}
    \label{fig:tmprss2}
\end{figure}
ACE2 protein also finds substantial mentionings in the published literature. 
We show here the subnetwork of ACE2 in Fig.~\ref{fig:ace2subnetwork}.
The subnetwork has all nodes connected to ACE2 and their interconnections.
It is a large network with 367 nodes and 1141 edges. 
There is a total of 60 drugs, 173 proteins, 90 diseases, and 44 organisms in the ACE2 network.
We did path analysis to uncover more lead compounds in the ACE2 network. We found all the paths between ACE2 and
Spike protein with a maximum of 3 hops between the two nodes. We further impose a condition that all nodes
in the path should either be a drug or a protein.
The following drugs or drug like compounds were found in the paths: A291P, Alanine, Arbidol, Chloroquine, Emodin,
Glutathione, HR2P-M2, IL-4, K267N, MAB 1a9, Nitric Oxide, Rabbit Antisera, Sialic acid,
 SP-10, SP-8, Superoxide, and TAPI-2.
 K267E and A291P are actually polymorphism to DDP4 host protease.
 DDP4 helps the binding of Spike protein to the host. It was observed that these polymorphism reduce viral replication and thus have a therapeutic effects\cite{kleine2020polymorphisms}. Arbidol is an antiviral drug that has been reported to block viral entry and replication\cite{pecheur2016synthetic}.
 Emodin is another top drug in the ACE2 network that works by blocking the interaction between ACE2 and Spike protein\cite{ho2007emodin}. Glutathione works by downregulating ACE2\cite{xu2018excessive}.
 SP-8 and SP-10 are peptides that disrupt the binding of Spike protein to ACE2\cite{ho2006design}.

While discussing ACE2, it is worth mentioning the role of TMPRSS2 in the pathogenesis.
TMPRSS2 is a serine protease that plays a role in cleaving the spike protein and helps the binding of S protein to ACE2.
Figure~\ref{fig:tmprss2} shows the TMPRSS2 subnetwork. It is a small and yet important subnetwork and is easy
to visualize. It has 37 nodes and 95 edges. There are 2 drugs, 20 proteins, 3 diseases, and 12 organisms in the network.
Camostat which is a TMPRSS2 inhibitor shows up the network and can be used as a drug to block the viral entry. One of the important protein that shows in the network in Fig.~\ref{fig:tmprss2} is LY6E.
It is an interferon stimulated protein and has been shown to be effective in curbing the entry
of SARS-CoV-2 in a couple of studies\cite{pfaender2020ly6e,zhao2020ly6e}.

\subsection{The Viral Protein Network}
\begin{figure}[h!]
    \centering
    \includegraphics[width=0.95\textwidth]{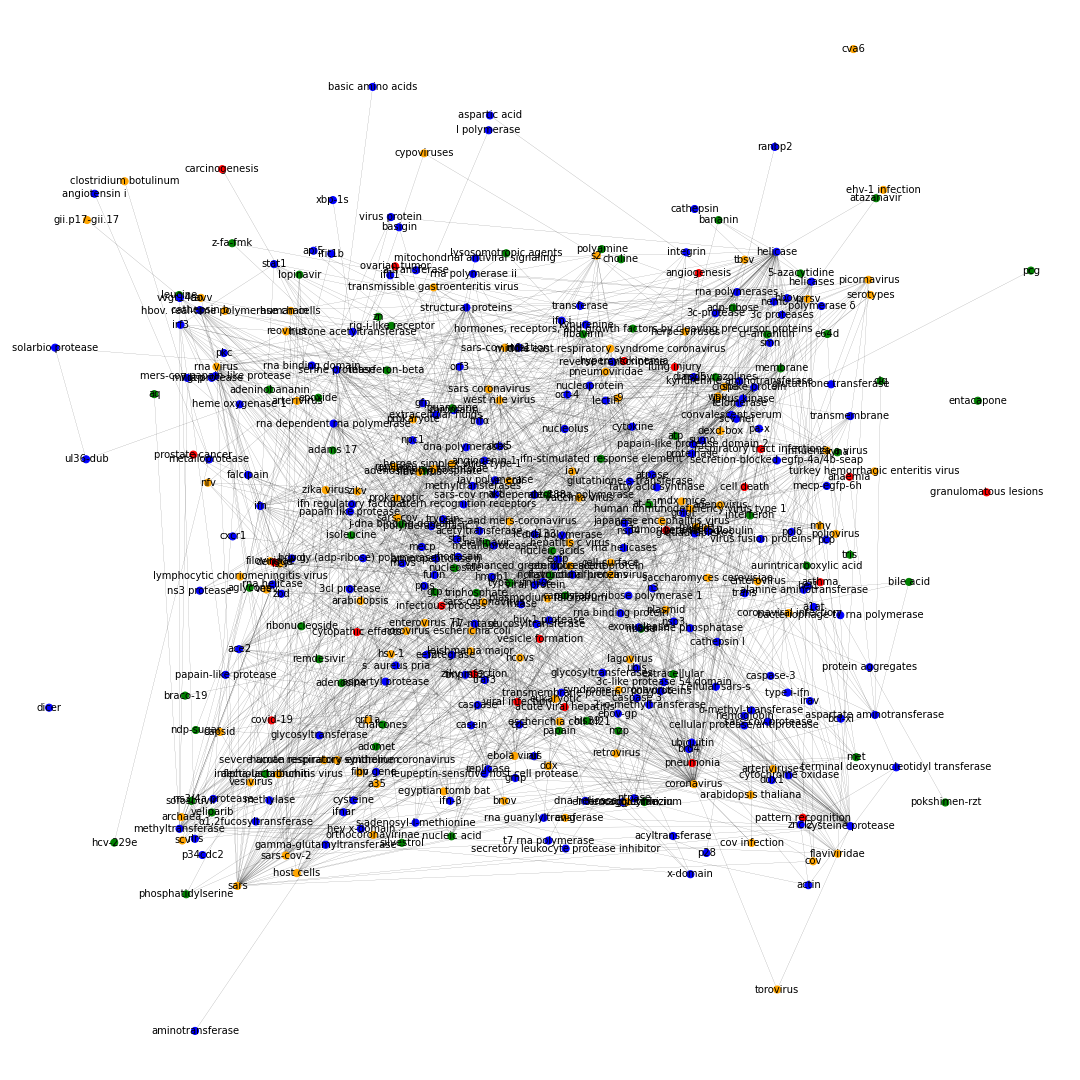}
    \caption{The subnetwork composed of polymerase, protease, helicase, transferase and related entities}
    \label{fig:targetnetwork}
\end{figure}
We shift our attention now to the viral target protein network. Figure \ref{fig:targetnetwork} shows
a subnetwork of concepts related to polymerase, protease, helicase, and transferase.
The RNA dependent RNA polymerase (RdRp) is one of the important viral targets.
RdRp is an enzyme that catalyzes the replication of RNA from an RNA template. Several drugs have been proposed over
the years to inhibit the function of RdRp. Remdesivir has emerged as one of the more promising drugs.
Remdesivir treatment is prohibitively expensive. However, several other drugs have emerged out of our literature mining.
In fact, one of the drugs with a higher centrality measure is Adenosine which is an Adenosine triphoshate analog and can successfully block the viral replication\cite{hercik2017adenosine}. 
There are few more drugs that have emerged targeting RdRp namely Sofosbuvir, AZT, Tenofovir Alafenamide, and Alovudine.
We also analyzed the 3C-like protease network and 3 experimental drugs named EPDTC, JMF1586, and JMF1600 emerged.

\subsection{The network based drug discovery}
\begin{figure}[h!]
    \centering
    \includegraphics[width=\textwidth]{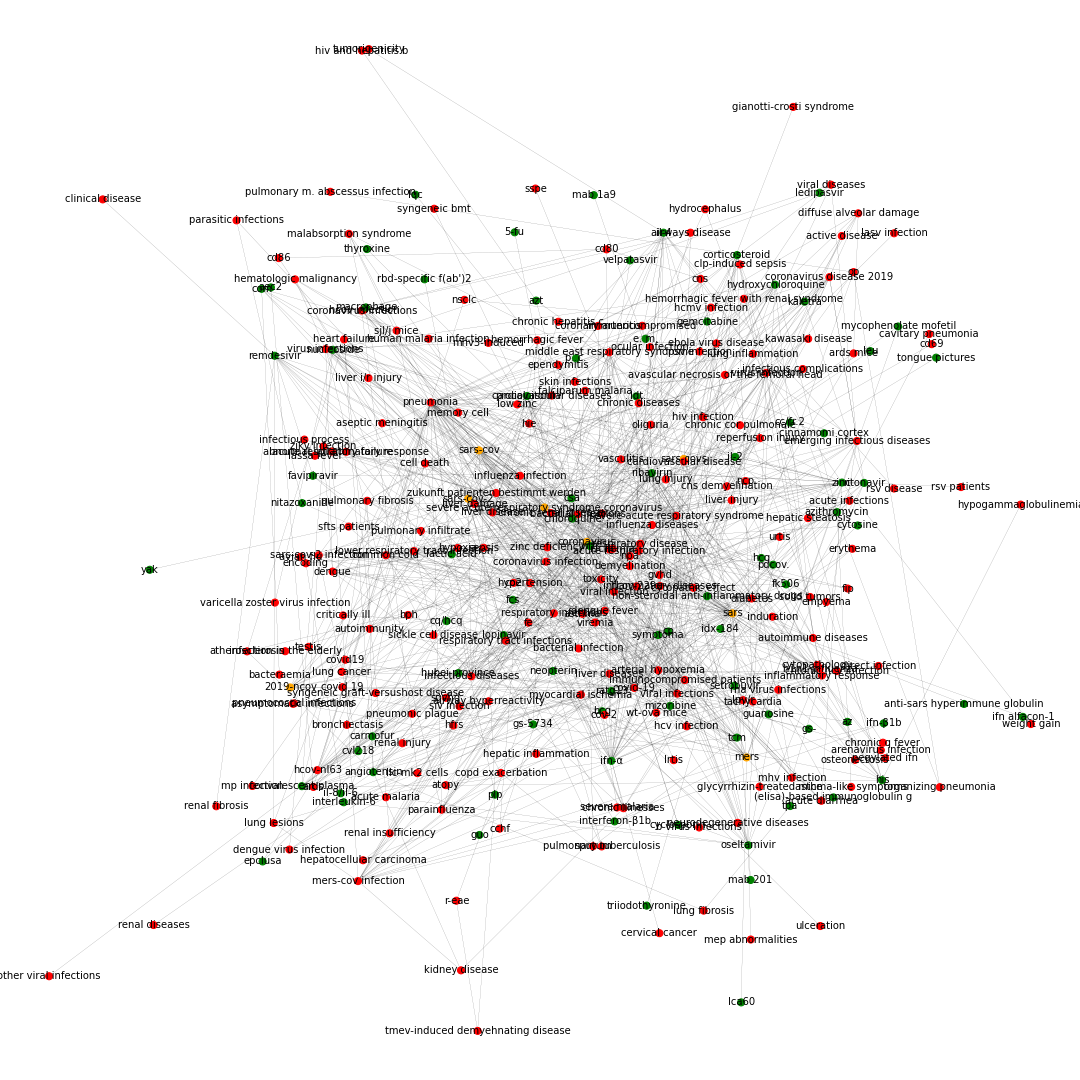}
    \caption{Drug disease network primary set of drugs related to COVID-19}
    \label{fig:drugnetwork}
\end{figure}
\begin{table}[h!]
    \centering
    \begin{tabular}{||c|c|c||}
\hline
Rank & Drug & Centrality Measure \\ [0.5ex] 
\hline\hline
1 & type i ifn & 0.0401 \\
\hline
2 & ifn-$\alpha$ & 0.0274 \\
\hline
3 & ribavirin & 0.0222 \\
\hline
4 & nitric oxide & 0.0198 \\
\hline
5 & sialic acid & 0.0180 \\
\hline
6 & mabs & 0.0177 \\
\hline
7 & il-4 & 0.0172 \\
\hline
8 & DCs & 0.0172 \\
\hline
9 & chloroquine & 0.0169 \\
\hline
10 & lipid & 0.0153 \\
\hline
11 & il-2 & 0.0152 \\
\hline
12 & ifn- & 0.0129 \\
\hline
13 & zinc & 0.0127 \\
\hline
14 & flavonoid & 0.0120 \\
\hline
15 & quercetin & 0.0119 \\
\hline
16 & atp & 0.0112 \\
\hline
17 & glycyrrhizin & 0.0112 \\
\hline
18 & angiotensin & 0.0107 \\
\hline
19 & emodin & 0.0106 \\
\hline
20 & dexamethasone & 0.0103 \\
\hline
    \end{tabular}
    \caption{Top drugs in the network}
    \label{tab:topdrug}
\end{table}
We also ranked all the drugs discovered by their centrality measure. The results are shown in Table \ref{tab:topdrug}.
The unlikely candidate here is Type I IFN or Type I Interferon. 
The second candidate is IFN-$\alpha$ which is related to Type I IFN. The third candidate is Ribavirin which is very popular
anitviral drug effective against many different species of virus. 
At first, these results surprised us.
However, some of the recent experiments have corroborated our findings.
The interferon signalling system plays a key role in immune system response to virus infestations.
It was observed in recent study the SARS-CoV-2 suppresses the induction of type I interferons
while aggravating IL-6 response.
These results suggest an important immune imbalance where low levels of interferons reduces the hosts ability to contain the viral replication, and the activation of IL-6 immune responses promotes inflammation\cite{Blanco-Melo2020.03.24.004655}.
In a recent phase 2 clinical trail, a recombinant type I IFN combined with Ribavirin and Lopinavir demonstrated good results in curing the COVID-19 ailment
compared to administering Ribavirin alone\cite{hung2020triple,shalhoubinterferon}.

We further tried to uncover the drug used for dealing with SARS-CoV-2 like viruses by interrogating our
network for following concepts: \textbf{coronavirus}, \textbf{SARS}, \textbf{SARS-CoV}, \textbf{SARS-CoV-2}, \textbf{COVID-19}, \textbf{2019-ncov}, \textbf{ARDS}, \textbf{MERS}, \textbf{MERS-CoV infection}, \textbf{MERS-CoV}, \textbf{RNA virus infections}, \textbf{SARS-CoV mouse lung}, \textbf{SARS-CoV particles}, \textbf{SARS-CoV populations}, \textbf{SARS-CoV-2 infection}, \textbf{Coronavirus infection}, \textbf{severe acute respiratory syndrome}, and, \textbf{severe acute respiratory syndrome coronavirus}.
We restricted the relationship type as being TREATS.
Table \ref{tab:drugdisease} in Appendix \ref{sec:appendix} shows these relations. We uncovered about 70 such relationships.
The hyperlink to the original article can be found in fifth column \emph{Reference Id}.
It is interesting to note that we discovered the a treatment that have been recently reported to be effective in COVID-19 and
have also been reported earlier in reference to another ailment.
A combination treatment with IFN-1b, Lopinavir/Ritonavir, and Ribavirin
which was recently reported to be successful in phase 2 clinical trial for COVID-19\cite{hung2020triple}.
We discovered that the same combination was proposed earlier for MERS (see entry 23)\cite{arabi2018treatment}.
Thus, the entries in Table \ref{tab:drugdisease} may serve as a ready reference point to explore many other treatment modalities for COVID-19.
The extensive drug-disease network corresponding to Table \ref{tab:drugdisease} is shown in Fig.~\ref{fig:drugnetwork}. This subnetwork was formed by taking all the drugs listed in Table \ref{tab:drugdisease} and finding all the diseases related to these drugs.

\section{Conclusion}
We undertook a comprehensive concept identification and network analysis for COVID-19.
We demonstrated the use of a novel concept recognition and relationship discovery
engine that crafts some of the latest advances in natural language processing into
a state-of-the-art solution for biomedical entity recognition and relationship discovery
problem. Several new drugs were uncovered through the studies and many different treatment modalities
were brought to the surface.
We envision these solutions to have a wide ranging impact through the length and breadth of drug discovery process spanning all therapeutic areas.

\newpage
\section{\label{sec:appendix}Appendix}
\begin{longtable}[c]{|| m{0.5cm} | m{8cm} | m{2.5cm} | m{2.5cm} | m{2.5cm} ||}
    \caption{Drug Disease Release Relationship\label{tab:drugdisease}}\\
    \hline
    & Sentence & Drug & Disease & Reference Id \\
\hline
\hline
1 & However, in our prediction, they may also bind to the replication complex components of 2019-nCoV with an inhibitory potency with Kd < 1000 nM. In addition, we also found that several antiviral agents, such as Kaletra, could be used for the treatment of 2019-nCoV, although there is no real-world evidence supporting the prediction. & Kaletra & 2019-nCoV & \href{https://pubmed.ncbi.nlm.nih.gov/32280433/}{32280433} \\
\hline
2 & Our observation suggests that AEC2 attenuates LPS-induced ARDS via the Ang-(1-7) Mas pathway by inhibiting ERK/NF-kB activation. & AEC2 & ARDS & \href{https://pubmed.ncbi.nlm.nih.gov/27302421/}{27302421} \\
\hline
3 & Methods: PubMed, EMBASE, and three trial Registries were searched for studies on the use of chloroquine in patients with COVID-19. & chloroquine & COVID-19 & \href{https://pubmed.ncbi.nlm.nih.gov/32173110/}{32173110} \\
\hline
4 & The aim of this systematic review was to summarize the evidence regarding chloroquine for the treatment of COVID-19. & chloroquine & COVID-19 & \href{https://pubmed.ncbi.nlm.nih.gov/32173110/}{32173110} \\
\hline
5 & Furthermore, integrated analysis predicted that IL-1$\beta$ and M-CSF may be novel candidate target genes for inflammatory storm and that TNFSF13, IL-18, IL-2 and IL-4 may be beneficial for the recovery of COVID-19 patients. & IL-4 & COVID-19 & \href{https://pubmed.ncbi.nlm.nih.gov/32377375/}{32377375} \\
\hline
6 & Furthermore, integrated analysis predicted that IL-1$\beta$ and M-CSF may be novel candidate target genes for inflammatory storm and that TNFSF13, IL-18, IL-2 and IL-4 may be beneficial for the recovery of COVID-19 patients. & IL-2 & COVID-19 & \href{https://pubmed.ncbi.nlm.nih.gov/32377375/}{32377375} \\
\hline
7 & An FDA-approved antineoplastic drug, 30 carmofur, has been identified as an inhibitor that targets COVID-19 virus M pro .However, its inhibitory mechanism is unknown. & carmofur & COVID-19 & \href{http://dx.doi.org/10.1038/s41594-020-0440-6}{10.1038} \\
\hline
8 & This study aims to evaluate the efficacy of hydroxychloroquine (HCQ) in the treatment of patients with COVID-19.Main methods: From February 4 to February 28, 2020, 62 patients suffering from COVID-19 were diagnosed and admitted to Renmin Hospital of Wuhan University. & HCQ & COVID-19 & \href{https://www.medrxiv.org/content/10.1101/2020.03.22.20040758v3}{20040758} \\
\hline
11 & Our study (NCT04252885), designated as ELACOI, was an exploratory randomized (2:2:1) and controlled one, exploring the efficacy and safety of lopinavir/ritonavir (LPV/r) or arbidol monotherapy treating mild/moderate COVID-19 patients. & lopinavir & COVID-19 & \href{https://europepmc.org/article/ppr/ppr118529}{118529} \\
\hline
12 & Hydroxychloroquine has recently received Emergency Use Authorization by the FDA and is currently prescribed in combination with azithromycin for COVID-19 pneumonia. & azithromycin & COVID-19 & \href{https://www.medrxiv.org/content/10.1101/2020.04.08.20054551v2}{20054551} \\
\hline
13 & Our data further supports the view that universal BCG vaccination has a protective effect on the course of COVID-19 probably preventing progression to severe disease and death. & BCG & COVID-19 & \href{https://www.medrxiv.org/content/10.1101/2020.04.07.20053272v2}{20053272} \\
\hline
14 & Herein, we examine the effects of Favipiravir (FPV) versus Lopinavir (LPV)/ritonavir (RTV) for the treatment of COVID-19. & Lopinavir & COVID-19 & \href{https://pubmed.ncbi.nlm.nih.gov/32346491/}{32346491} \\
\hline
15 & Herein, we examine the effects of Favipiravir (FPV) versus Lopinavir (LPV)/ritonavir (RTV) for the treatment of COVID-19. & Favipiravir & COVID-19 & \href{https://pubmed.ncbi.nlm.nih.gov/32346491/}{32346491} \\
\hline
16 & Remdesivir (GS-5734™) is a broad-spectrum antiviral drug that is now being tested as a potential treatment for COVID-19 in international, multi-site clinical trials. & Remdesivir & COVID-19 & \href{https://pubmed.ncbi.nlm.nih.gov/32258351/}{32258351} \\
\hline
17 & We aimed to evaluate the definite efficacy and safety of corticosteroid in the treatment of severe COVID-19 pneumonia. & corticosteroid & COVID-19 & \href{https://www.medrxiv.org/content/10.1101/2020.03.06.20032342v1}{20032342} \\
\hline
18 & We also discussed several putative mechanisms of the anti-SARS-CoV-2 effects for CVL218 or other PARP1 inhibitors to be involved in the treatment of COVID-19. & CVL218 & COVID-19 & \href{https://europepmc.org/article/ppr/ppr117314}{117314} \\
\hline
19 & In summary, the PARP1 inhibitor CVL218 discovered by our data-driven drug repositioning framework can serve as a potential therapeutic agent for the treatment of COVID-19. & CVL218 & COVID-19 & \href{https://europepmc.org/article/ppr/ppr117314}{117314} \\
\hline
20 & Our in silico screening followed by wet-lab validation indicated that a poly-ADP-ribose polymerase 1 (PARP1) inhibitor, CVL218, currently in Phase I clinical trial, may be repurposed to treat COVID-19. & CVL218 & COVID-19 & \href{https://europepmc.org/article/ppr/ppr117314}{117314} \\
\hline
21 & We aimed to compare arbidol and lopinavir/ritonavir(LPV/r) treatment for patients with COVID-19 with LPV/r only. & ritonavir & COVID-19 & \href{https://pubmed.ncbi.nlm.nih.gov/32171872/}{32171872} \\
\hline
22 & We aimed to compare arbidol and lopinavir/ritonavir(LPV/r) treatment for patients with COVID-19 with LPV/r only. & lopinavir & COVID-19 & \href{https://pubmed.ncbi.nlm.nih.gov/32171872/}{32171872} \\
\hline
23 & The MIRACLE trial (MERS-CoV Infection tReated with A Combination of Lopinavir/ritonavir and intErferon-$\beta$1b) investigates the efficacy of a combination therapy of lopinavir/ritonavir and recombinant interferon-$\beta$1b provided with standard supportive care, compared to placebo provided with standard supportive care, in hospitalized patients with laboratory-confirmed MERS. & interferon-$\beta$1b & MERS & \href{https://pubmed.ncbi.nlm.nih.gov/29382391/}{29382391} \\
\hline
24 & Results from in vitro and animal studies suggest that a combination of lopinavir/ritonavir and interferon-$\beta$1b (IFN-$\beta$1b) may be effective against MERS-CoV. The aim of this study is to investigate the efficacy of treatment with a combination of lopinavir/ritonavir and recombinant IFN-$\beta$1b provided with standard supportive care, compared to treatment with placebo provided with standard supportive care in patients with laboratory-confirmed MERS requiring hospital admission. & interferon-$\beta$1b & MERS & \href{https://pubmed.ncbi.nlm.nih.gov/29382391/}{29382391} \\
\hline
25 & Results from in vitro and animal studies suggest that a combination of lopinavir/ritonavir and interferon-$\beta$1b (IFN-$\beta$1b) may be effective against MERS-CoV. The aim of this study is to investigate the efficacy of treatment with a combination of lopinavir/ritonavir and recombinant IFN-$\beta$1b provided with standard supportive care, compared to treatment with placebo provided with standard supportive care in patients with laboratory-confirmed MERS requiring hospital admission. & IFN-$\beta$1b & MERS & \href{https://pubmed.ncbi.nlm.nih.gov/29382391/}{29382391} \\
\hline
26 & The data presented here support testing of the efficacy of remdesivir treatment in the context of a MERS clinical trial. & remdesivir & MERS & \href{https://pubmed.ncbi.nlm.nih.gov/32054787/}{32054787} \\
\hline
27 & In this study, we describe a method for the swift generation of a humanderived monoclonal antibody, known as LCA60, as a treatment for MERS infections. & LCA60 & MERS & \href{https://pubmed.ncbi.nlm.nih.gov/27102927/}{27102927} \\
\hline
28 & We assessed 3 repurposed drugs with potent in vitro anti-MERS-CoV activity (mycophenolate mofetil [MMF], lopinavir/ritonavir, and interferon-$\beta$1b) in common marmosets with severe disease resembling MERS in humans. & ritonavir & MERS & \href{https://pubmed.ncbi.nlm.nih.gov/26198719/}{26198719} \\
\hline
29 & We assessed 3 repurposed drugs with potent in vitro anti-MERS-CoV activity (mycophenolate mofetil [MMF], lopinavir/ritonavir, and interferon-$\beta$1b) in common marmosets with severe disease resembling MERS in humans. & mycophenolate mofetil & MERS & \href{https://pubmed.ncbi.nlm.nih.gov/26198719/}{26198719} \\
\hline
30 & We aimed to compare ribavirin and interferon alfa-2a treatment for patients with severe MERS-CoV infection with a supportive therapy only. & interferon alfa-2a & MERS-CoV infection & \href{https://pubmed.ncbi.nlm.nih.gov/25278221/}{25278221} \\
\hline
31 & We aimed to compare ribavirin and interferon alfa-2a treatment for patients with severe MERS-CoV infection with a supportive therapy only. & ribavirin & MERS-CoV infection & \href{https://pubmed.ncbi.nlm.nih.gov/25278221/}{25278221} \\
\hline
32 & These data indicate that saracatinib alone or in combination with gemcitabine can provide a new therapeutic option for the treatment of MERS-CoV infection. & saracatinib & MERS-CoV infection & \href{https://pubmed.ncbi.nlm.nih.gov/29795047/}{29785047} \\
\hline
33 & These data indicate that saracatinib alone or in combination with gemcitabine can provide a new therapeutic option for the treatment of MERS-CoV infection. & gemcitabine & MERS-CoV infection & \href{https://pubmed.ncbi.nlm.nih.gov/29795047/}{29785047} \\
\hline
34 & Having been used extensively in clinical trials and in post-marketing experience, nitazoxanide is an attractive drug candidate for treatment of Middle East respiratory syndrome. & nitazoxanide & Middle East respiratory syndrome & \href{https://pubmed.ncbi.nlm.nih.gov/27095301/}{27095301} \\
\hline
35 & The objective of this study was to evaluate the effect of ribavirin and recombinant interferon (RBV/rIFN) therapy on the outcomes of critically ill patients with Middle East respiratory syndrome (MERS), accounting for time-varying confounders. & recombinant interferon & Middle East respiratory syndrome & \href{https://pubmed.ncbi.nlm.nih.gov/31925415/}{31925415} \\
\hline
36 & The objective of this study was to evaluate the effect of ribavirin and recombinant interferon (RBV/rIFN) therapy on the outcomes of critically ill patients with Middle East respiratory syndrome (MERS), accounting for time-varying confounders. & ribavirin & Middle East respiratory syndrome & \href{https://pubmed.ncbi.nlm.nih.gov/31925415/}{31925415} \\
\hline
37 & Ribavirin is currently used for the treatment of several RNA virus infections clinically, so its anti-EV71 efficacy was evaluated. & Ribavirin & RNA virus infections & \href{https://pubmed.ncbi.nlm.nih.gov/18279075/}{18279075} \\
\hline
38 & A pilot clinical report showed effectiveness of IFN-$\alpha$ for the treatment of SARS patients. & IFN-$\alpha$ & SARS & \href{https://pubmed.ncbi.nlm.nih.gov/15174965/}{15174965} \\
\hline
39 & Results from a pilot study to evaluate the clinical efficacy of IFN-$\alpha$ treatment of SARS patients provided evidence for IFN-inducible resolution of disease. & IFN-$\alpha$ & SARS & \href{https://pubmed.ncbi.nlm.nih.gov/16474437/}{16474437} \\
\hline
40 & We therefore suggest that pegylated IFN-$\alpha$ protects type 1 pneumocytes from SCV infection, and should be considered a candidate drug for SARS therapy SARS has recently emerged in the human population as a potentially fatal respiratory disease. & IFN-$\alpha$ & SARS & \href{https://pubmed.ncbi.nlm.nih.gov/14981511/}{14981511} \\
\hline
41 & The efficacy of MAb 201 in the treatment of SARS was evaluated in golden Syrian hamsters, an animal model that supports SARS-CoV replication to high levels and displays severe pathological changes associated with infection, including pneumonitis and pulmonary consolidation. & MAb 201 & SARS & \href{https://pubmed.ncbi.nlm.nih.gov/16453264/}{16453264} \\
\hline
42 & The highest leukocyte and neutrophil counts, lactate dehydrogenase, and creatine kinase; positive endexpiratory pressure; and use of corticosteroids, ribavirin, and intravenous immunoglobulin were higher in the SARS group. & ribavirin & SARS & \href{https://pubmed.ncbi.nlm.nih.gov/19056023/}{190556023} \\
\hline
43 & Developing high titers of anti-SARS hyperimmune globulin to provide an alternative pathway for emergent future prevention and treatment of SARS. & anti-SARS hyperimmune globulin & SARS & \href{https://pubmed.ncbi.nlm.nih.gov/16297347/}{16297347} \\
\hline
44 & Based on these findings, a pilot study to evaluate the potential clinical benefit and safety of IFN alfacon-1 in SARS treatment was conducted by our laboratory. & IFN alfacon-1 & SARS & \href{https://www.nature.com/articles/7310030}{7310030} \\
\hline
45 & Cholesterol depletion by pretreatment of Vero E6 cells with methyl-b-cyclodextrin (MbCD) inhibited the production of SARS-CoV particles released from the infected cells. & methyl-b-cyclodextrin & SARS-CoV particles & \href{https://pubmed.ncbi.nlm.nih.gov/17194611/}{17194611} \\
\hline
46 & Comparison of fullgenome next-generation sequencing of 5-FU treated SARS-CoV populations revealed a 16-fold increase in the number of mutations within the ExoN2 population as compared to ExoN+. & 5-FU & SARS-CoV populations & \href{https://pubmed.ncbi.nlm.nih.gov/23966862/}{23966862} \\
\hline
47 & In addition, the results suggest guanosine derivative (IDX-184), Setrobuvir, and YAK as top seeds for antiviral treatments with high potential to fight the SARS-CoV-2 strain specifically. & guanosine & SARS-CoV-2 & \href{https://pubmed.ncbi.nlm.nih.gov/32222463/}{32222463} \\
\hline
48 & In addition, the results suggest guanosine derivative (IDX-184), Setrobuvir, and YAK as top seeds for antiviral treatments with high potential to fight the SARS-CoV-2 strain specifically. & IDX-184 & SARS-CoV-2 & \href{https://pubmed.ncbi.nlm.nih.gov/32222463/}{32222463} \\
\hline
49 & In addition, the results suggest guanosine derivative (IDX-184), Setrobuvir, and YAK as top seeds for antiviral treatments with high potential to fight the SARS-CoV-2 strain specifically. & Setrobuvir & SARS-CoV-2 & \href{https://pubmed.ncbi.nlm.nih.gov/32222463/}{32222463} \\
\hline
50 & In addition, the results suggest guanosine derivative (IDX-184), Setrobuvir, and YAK as top seeds for antiviral treatments with high potential to fight the SARS-CoV-2 strain specifically. & YAK & SARS-CoV-2 & \href{https://pubmed.ncbi.nlm.nih.gov/32222463/}{32222463} \\
\hline
51 & Neutralization test demonstrated that RBD-specific F(ab')2 inhibited SARS-CoV-2 with EC50 at 0.07 $\mu$g/ml, showing a potent inhibitory effect on SARS-CoV-2. & RBD-specific F(ab')2 & SARS-CoV-2 & \href{https://europepmc.org/article/ppr/ppr149893}{149893} \\
\hline
52 & Chloroquine has been sporadically used in treating SARS-CoV-2 infection. & Chloroquine & SARS-CoV-2 infection & \href{https://pubmed.ncbi.nlm.nih.gov/32150618/}{32150618} \\
\hline
53 & Treatment with convalescent plasma for critically ill patients with SARS-CoV-2 infection & convalescent plasma & SARS-CoV-2 infection & \href{https://pubmed.ncbi.nlm.nih.gov/32243945/}{32243945} \\
\hline
54 & MAb 1A9 is a broadly neutralizing mAb that prevents viral entry mediated by the S proteins of human and civet SARS-CoVs as well as bat SL-CoVs. & MAb 1A9 & SARS-CoVs & \href{https://pubmed.ncbi.nlm.nih.gov/25019613/}{25019613} \\
\hline
55 & Among these, the antivirals ledipasvir or velpatasvir are particularly attractive as therapeutics to combat the new coronavirus with minimal side effects, commonly fatigue and headache. & ledipasvir & coronavirus & \href{https://pubmed.ncbi.nlm.nih.gov/32194944/}{32194944} \\
\hline
56 & Among these, the antivirals ledipasvir or velpatasvir are particularly attractive as therapeutics to combat the new coronavirus with minimal side effects, commonly fatigue and headache. & velpatasvir & coronavirus & \href{https://pubmed.ncbi.nlm.nih.gov/32194944/}{32194944} \\
\hline
57 & For a random sample of 127 calves, serum zinc concentrations before and after treatment and a fecal antigen ELISA at diarrhea start and resolution for Escherichia coli K99, rotavirus, coronavirus, and Cryptosporidium parvum were performed. & zinc & coronavirus & \href{https://pubmed.ncbi.nlm.nih.gov/31291305/}{31291305} \\
\hline
58 & Based on our analysis of hepatitis C virus and coronavirus replication, and the molecular structures and activities of viral inhibitors, we previously reasoned that the FDA-approved heptatitis C drug EPCLUSA (Sofosbuvir/Velpatasvir) should inhibit coronaviruses, including SARS-CoV-2. & EPCLUSA & coronaviruses & \href{https://www.biorxiv.org/content/10.1101/2020.03.12.989186v1.full}{989186} \\
\hline
59 & Previously, in vitro studies have shown that cyclosporin (CsA) and FK506 inhibit virus replication in diverse coronaviruses. & FK506 & coronaviruses & \href{https://pubmed.ncbi.nlm.nih.gov/26675666/}{26675666} \\
\hline
60 & Previously, in vitro studies have shown that cyclosporin (CsA) and FK506 inhibit virus replication in diverse coronaviruses. & cyclosporin & coronaviruses & \href{https://pubmed.ncbi.nlm.nih.gov/26675666/}{26675666} \\
\hline
61 & We investigated the effect of massive doses of corticosteroid therapy on bone metabolism using specific biochemical markers of bone metabolism, and the prevalence of osteonecrosis in severe acute respiratory syndrome (SARS) patients at a university teaching hospital in Hong Kong. & corticosteroid & severe acute respiratory syndrome & \href{https://pubmed.ncbi.nlm.nih.gov/16753744/}{16753744} \\
\hline
62 & Ribavirin and corticosteroids were used widely as front-line treatments for severe acute respiratory syndrome; however, previous evaluations were inconclusive. & Ribavirin & severe acute respiratory syndrome & \href{https://pubmed.ncbi.nlm.nih.gov/19958895/}{19958895} \\
\hline
63 & We assessed the effectiveness of ribavirin and corticosteroids as the initial treatment for severe acute respiratory syndrome using propensity score analysis. & ribavirin & severe acute respiratory syndrome & \href{https://pubmed.ncbi.nlm.nih.gov/19958895/}{19958895} \\
\hline
64 & In order to evaluate the efficacy of convalescent plasma therapy in the treatment of patients with severe acute respiratory syndrome (SARS), 80 SARS patients were given convalescent plasma at Prince of Wales Hospital, Hong Kong, between 20 March and 26 May 2003. & convalescent plasma & severe acute respiratory syndrome & \href{https://pubmed.ncbi.nlm.nih.gov/15616839/}{15616839} \\
\hline
65 & A new enzyme-linked immunosorbent assay (ELISA)-based immunoglobulin G (IgG)-plus-IgM antibody detection test for severe acute respiratory syndrome (SARS) has been developed by using a cocktail of four recombinant polypeptides as the antigen. & (ELISA)-based immunoglobulin G & severe acute respiratory syndrome & \href{https://pubmed.ncbi.nlm.nih.gov/19038782/}{19038782} \\
\hline
66 & Lianhua-Qingwen capsule (LQC) is a commonly used Chinese medical preparation to treat viral influenza and especially played a very important role in the fight against severe acute respiratory syndrome (SARS) in 2002-2003 in China. & LQC & severe acute respiratory syndrome & \href{https://pubmed.ncbi.nlm.nih.gov/25654135/}{25654135} \\
\hline
67 & In this study, we used MRI to observe the lesion size changes of ONFH induced by corticosteroid administration in severe acute respiratory syndrome (SARS) patients. & corticosteroid & severe acute respiratory syndrome & \href{https://pubmed.ncbi.nlm.nih.gov/19533123/}{19533123} \\
\hline
68 & Study objective: To investigate the efficacy and safety profiles of corticosteroid therapy in severe acute respiratory syndrome (SARS) patients. & corticosteroid & severe acute respiratory syndrome
& \href{https://pubmed.ncbi.nlm.nih.gov/16778260/}{16778260} \\
\hline
69 & We found that the butanol fraction of Cinnamomi Cortex (CC/Fr.2) showed moderate inhibitory activity in wild-type severe acute respiratory syndrome coronavirus (wtSARS-CoV) and HIV/SARS-CoV S pseudovirus infections. & Cinnamomi Cortex & severe acute respiratory syndrome coronavirus & \href{https://pubmed.ncbi.nlm.nih.gov/19428598/}{19428598}\\
\hline
70 & We recently reported that the nucleoside analogue GS-5734 (remdesivir) potently inhibits human and zoonotic CoVs in vitro and in a severe acute respiratory syndrome coronavirus (SARS-CoV) mouse model. & GS-5734 & severe acute respiratory syndrome coronavirus & \href{https://europepmc.org/article/pmc/pmc7102703}{7102703} \\
\hline
\end{longtable}


\begin{thebibliography}{10}
\urlstyle{rm}
\expandafter\ifx\csname url\endcsname\relax
  \def\url#1{\texttt{#1}}\fi
\expandafter\ifx\csname urlprefix\endcsname\relax\def\urlprefix{URL }\fi
\expandafter\ifx\csname doiprefix\endcsname\relax\def\doiprefix{DOI: }\fi
\providecommand{\bibinfo}[2]{#2}
\providecommand{\eprint}[2][]{\url{#2}}

\bibitem{nichkawde1}
\bibinfo{author}{Singla, P.}, \bibinfo{author}{Singh, K.},
  \bibinfo{author}{Sarode, K.}, \bibinfo{author}{Chandrakar, A.} \&
  \bibinfo{author}{Nichkawde, C.}
\newblock \bibinfo{journal}{\bibinfo{title}{Biomedical entity recognition using
  a masked concept model}}.
\newblock {\emph{\JournalTitle{manuscript under preparation}}}
  (\bibinfo{year}{2020}).

\bibitem{vaswani2017attention}
\bibinfo{author}{Vaswani, A.} \emph{et~al.}
\newblock \bibinfo{title}{Attention is all you need}.
\newblock In \emph{\bibinfo{booktitle}{Advances in neural information
  processing systems}}, \bibinfo{pages}{5998--6008} (\bibinfo{year}{2017}).

\bibitem{nichkawde2}
\bibinfo{author}{Singh, K.}, \bibinfo{author}{Singla, P.},
  \bibinfo{author}{Sarode, K.}, \bibinfo{author}{Chandrakar, A.} \&
  \bibinfo{author}{Nichkawde, C.}
\newblock \bibinfo{journal}{\bibinfo{title}{Biomedical relationship discovery
  using a masked concept model}}.
\newblock {\emph{\JournalTitle{manuscript under preparation}}}
  (\bibinfo{year}{2020}).

\bibitem{wang2020cord}
\bibinfo{author}{Wang, L.~L.} \emph{et~al.}
\newblock \bibinfo{journal}{\bibinfo{title}{Cord-19: The covid-19 open research
  dataset}}.
\newblock {\emph{\JournalTitle{ArXiv}}}  (\bibinfo{year}{2020}).

\bibitem{katz1953new}
\bibinfo{author}{Katz, L.}
\newblock \bibinfo{journal}{\bibinfo{title}{A new status index derived from
  sociometric analysis}}.
\newblock {\emph{\JournalTitle{Psychometrika}}} \textbf{\bibinfo{volume}{18}},
  \bibinfo{pages}{39--43} (\bibinfo{year}{1953}).

\bibitem{gordon2020sars}
\bibinfo{author}{Gordon, D.~E.} \emph{et~al.}
\newblock \bibinfo{journal}{\bibinfo{title}{A sars-cov-2 protein interaction
  map reveals targets for drug repurposing}}.
\newblock {\emph{\JournalTitle{Nature}}} \bibinfo{pages}{1--13}
  (\bibinfo{year}{2020}).

\bibitem{kleine2020polymorphisms}
\bibinfo{author}{Kleine-Weber, H.} \emph{et~al.}
\newblock \bibinfo{journal}{\bibinfo{title}{Polymorphisms in dipeptidyl
  peptidase 4 reduce host cell entry of middle east respiratory syndrome
  coronavirus}}.
\newblock {\emph{\JournalTitle{Emerging microbes \& infections}}}
  \textbf{\bibinfo{volume}{9}}, \bibinfo{pages}{155--168}
  (\bibinfo{year}{2020}).

\bibitem{pecheur2016synthetic}
\bibinfo{author}{P{\'e}cheur, E.-I.} \emph{et~al.}
\newblock \bibinfo{journal}{\bibinfo{title}{The synthetic antiviral drug
  arbidol inhibits globally prevalent pathogenic viruses}}.
\newblock {\emph{\JournalTitle{Journal of virology}}}
  \textbf{\bibinfo{volume}{90}}, \bibinfo{pages}{3086--3092}
  (\bibinfo{year}{2016}).

\bibitem{ho2007emodin}
\bibinfo{author}{Ho, T.-Y.}, \bibinfo{author}{Wu, S.-L.},
  \bibinfo{author}{Chen, J.-C.}, \bibinfo{author}{Li, C.-C.} \&
  \bibinfo{author}{Hsiang, C.-Y.}
\newblock \bibinfo{journal}{\bibinfo{title}{Emodin blocks the sars coronavirus
  spike protein and angiotensin-converting enzyme 2 interaction}}.
\newblock {\emph{\JournalTitle{Antiviral research}}}
  \textbf{\bibinfo{volume}{74}}, \bibinfo{pages}{92--101}
  (\bibinfo{year}{2007}).

\bibitem{xu2018excessive}
\bibinfo{author}{Xu, J.}, \bibinfo{author}{Sriramula, S.} \&
  \bibinfo{author}{Lazartigues, E.}
\newblock \bibinfo{journal}{\bibinfo{title}{Excessive glutamate stimulation
  impairs ace2 activity through adam17-mediated shedding in cultured cortical
  neurons}}.
\newblock {\emph{\JournalTitle{Cellular and molecular neurobiology}}}
  \textbf{\bibinfo{volume}{38}}, \bibinfo{pages}{1235--1243}
  (\bibinfo{year}{2018}).

\bibitem{ho2006design}
\bibinfo{author}{Ho, T.-Y.} \emph{et~al.}
\newblock \bibinfo{journal}{\bibinfo{title}{Design and biological activities of
  novel inhibitory peptides for sars-cov spike protein and
  angiotensin-converting enzyme 2 interaction}}.
\newblock {\emph{\JournalTitle{Antiviral research}}}
  \textbf{\bibinfo{volume}{69}}, \bibinfo{pages}{70--76}
  (\bibinfo{year}{2006}).

\bibitem{pfaender2020ly6e}
\bibinfo{author}{Pfaender, S.} \emph{et~al.}
\newblock \bibinfo{journal}{\bibinfo{title}{Ly6e impairs coronavirus fusion and
  confers immune control of viral disease}}.
\newblock {\emph{\JournalTitle{bioRxiv}}}  (\bibinfo{year}{2020}).

\bibitem{zhao2020ly6e}
\bibinfo{author}{Zhao, X.} \emph{et~al.}
\newblock \bibinfo{journal}{\bibinfo{title}{Ly6e restricts the entry of human
  coronaviruses, including the currently pandemic sars-cov-2}}.
\newblock {\emph{\JournalTitle{bioRxiv}}}  (\bibinfo{year}{2020}).

\bibitem{hercik2017adenosine}
\bibinfo{author}{Herc{\'\i}k, K.} \emph{et~al.}
\newblock \bibinfo{journal}{\bibinfo{title}{Adenosine triphosphate analogs can
  efficiently inhibit the zika virus rna-dependent rna polymerase}}.
\newblock {\emph{\JournalTitle{Antiviral research}}}
  \textbf{\bibinfo{volume}{137}}, \bibinfo{pages}{131--133}
  (\bibinfo{year}{2017}).

\bibitem{Blanco-Melo2020.03.24.004655}
\bibinfo{author}{Blanco-Melo, D.} \emph{et~al.}
\newblock \bibinfo{journal}{\bibinfo{title}{Sars-cov-2 launches a unique
  transcriptional signature from in vitro, ex vivo, and in vivo systems}}.
\newblock {\emph{\JournalTitle{bioRxiv}}}
  \doiprefix\url{10.1101/2020.03.24.004655} (\bibinfo{year}{2020}).
\newblock
  \eprint{https://www.biorxiv.org/content/early/2020/03/24/2020.03.24.004655.full.pdf}.

\bibitem{hung2020triple}
\bibinfo{author}{Hung, I. F.-N.} \emph{et~al.}
\newblock \bibinfo{journal}{\bibinfo{title}{Triple combination of interferon
  beta-1b, lopinavir--ritonavir, and ribavirin in the treatment of patients
  admitted to hospital with covid-19: an open-label, randomised, phase 2
  trial}}.
\newblock {\emph{\JournalTitle{The Lancet}}}  (\bibinfo{year}{2020}).

\bibitem{shalhoubinterferon}
\bibinfo{author}{Shalhoub, S.}
\newblock \bibinfo{journal}{\bibinfo{title}{Interferon beta-1b for covid-19}}.
\newblock {\emph{\JournalTitle{The Lancet}}} .

\bibitem{arabi2018treatment}
\bibinfo{author}{Arabi, Y.~M.} \emph{et~al.}
\newblock \bibinfo{journal}{\bibinfo{title}{Treatment of middle east
  respiratory syndrome with a combination of lopinavir-ritonavir and
  interferon-$\beta$1b (miracle trial): study protocol for a randomized
  controlled trial}}.
\newblock {\emph{\JournalTitle{Trials}}} \textbf{\bibinfo{volume}{19}},
  \bibinfo{pages}{81} (\bibinfo{year}{2018}).

\end{thebibliography}
\end{document}